\pdfoutput=1

\documentclass[11pt]{article}

\usepackage[]{acl}

\usepackage{times}
\usepackage{latexsym}

\usepackage{graphicx}
\usepackage{hyperref}
\usepackage{amsmath}
\usepackage{multirow}
\usepackage{hhline}

\usepackage[T1]{fontenc}
\usepackage[utf8]{inputenc}

\usepackage{microtype}

\title{Learning Functional Distributional Semantics with Visual Data}

\author{Yinhong Liu, Guy Emerson \\
Department of Computer Science and Technology\\
University of Cambridge\\
  \texttt{\{yl535, gete2\}@cam.ac.uk} \\}

\date{}

\begin{document}
\maketitle
\begin{abstract}
Functional Distributional Semantics is a recently proposed framework for learning distributional semantics that provides linguistic interpretability. 
It models the meaning of a word as a binary classifier rather than a numerical vector. 
In this work, we propose a method to train a Functional Distributional Semantics model with grounded visual data.
We train it on the Visual Genome dataset,
which is closer to the kind of data encountered in human language acquisition than a large text corpus.
On four external evaluation datasets,
our model outperforms previous work on learning semantics from Visual Genome.
\footnote{Our code and models are publicly available at: \url{https://github.com/williamLyh/PixieVGModel}}


\end{abstract}

\section{Introduction}

The target of distributional semantics models is to understand and represent the meanings of words from their distributions in large corpus. 
Many approaches learn a numerical vector for each word, which encodes its distributional information. 
They can be roughly divided into two categories: frequency-based methods such as co-occurrence matrix \citep{Sahlgren2006TheWM}, and prediction-based methods such as Word2vec \citep{mikolov2013efficient}.
More recently, progress has been made in learning word representations in a specific context, which are also called contextualized embeddings. Examples include ELMo \citep{peters-etal-2018-deep} and BERT \citep{devlin-etal-2019-bert}.

Functional Distributional Semantics is a framework that not only provides contextualized semantic representations,
but also provides more interpretability.
It was first proposed by \citet{emerson-copestake-2016-functional},
and it explicitly separates the modeling of words and the modeling of objects and events. 
This is a fundamental distinction in predicate logic.
While logic is not necessary for all NLP tasks,
it is an essential tool for modeling many semantic phenomena
(for example, see: \citealp{cann1993sem,allan2001sem,kamp2013sem}).
For semantic research questions, having a logical interpretation is a clear advantage over vector-based models.
We will explain the framework in Section~\ref{sec:fds}.

Another issue with distributional semantic models, as discussed by \citet{emerson-2020-goals}, is the symbol grounding problem -- if meanings of words are defined in terms of other words, the definitions are circular. 
During human language acquisition, words are learned while interacting with the physical world, rather than from text or speech alone.
An important goal for a semantic theory is to explain how language relates to the world,
and how this relationship is learned.
We focus on the Visual Genome dataset,
not only because it provides relatively fine-grained annotations, but also it is similar to realistic circumstance encountered during language acquisition, as we will explain in Section~\ref{sec:vg}.


Our main theoretical contribution is to adapt the Functional Distributional Semantics framework to better suit visual data.
This is a step approaching the completion of long-term goal: leveraging previous work \citep{emerson2020autoencoding}, we could joint train the Functional Distributional Semantics model with both textual and visual data.
In order to make it compatible with modern techniques for machine vision,
while retaining its logical interpretability,
we replace the RBM of previous work with a Gaussian MRF,
as explained in Section~\ref{sec:model}.

Our main empirical contribution is to demonstrate the effectiveness of the resulting model.
In Section~\ref{sec:intrinsic}, we intrinsically evaluate the major components of our model, to see how well they fit the training data.
In Section~\ref{sec:extrinsic}, we evaluate our model on four external evaluation datasets, comparing against previous approaches to learning from Visual Genome, as well as strong text-based baselines.
Not only do we confirm \citet{herbelot-2020-solve}'s finding that learning from grounded data is more data-efficient than learning from text alone,
but our model outperforms the previous approaches, demonstrating the value of our functional approach.

\section{Background and Related Work}
\subsection{Visually Grounded Semantic Learning}
There is extensive research on learning language semantics from grounded visual data.
Visual-Semantic Embedding and Visual Concept Learning in Visual Question Answering are two representative frameworks. 
Some works under these frameworks share the idea with our Functional Distributional Semantics model that textual labels are modeled as classifiers over the semantic space.


Visual-Semantic Embedding \citep{frome2013devise} learns joint representations of vision and language in a common visual-semantic space. \citet{kiros2014unifying} proposed to unify the textual and visual embeddings via multimodal neural-based language models. 
\citet{ren2016joint} models images as points in the Visual-Semantic space, while text are Gaussian distributions over them.

Visual Concept Learning contributes to various visual linguistic applications, such as image captioning \citep{karpathy2015deep} and Visual Question Answering \citep{antol2015vqa}.
Some works in applying neural symbolic approach to VQA share similar ideas of learning visual concepts with our model.
For example, \citet{mao2018neuro} learn neural operators to capture attributes (concepts) of objects and map them into attribute-specific space. Then questions are parsed into executable programs. 
\citet{han2019visual} further learn the relations between objects as metaconcepts.

Our work differs from them in two main aspects.
Firstly, our framework supports truth-conditional semantics, as explained in Section~\ref{sec:fds}, and therefore provides more logical interpretability. 
Unlike the above works which always assume images are given, we use a generative model which allows us to perform inference on textual labels alone, as illustrated in Fig.~\ref{infergraph} and explained in Section~\ref{VI}.
Secondly, we learn semantics from the Visual Genome dataset, which is considered more similar to the data encountered during language acquisition, as explained in Section~\ref{sec:vg}.



\subsection{Functional Distributional Semantics}
\label{sec:fds}
Functional Distributional Semantics
was first proposed by \citet{emerson-copestake-2016-functional}.
The framework takes model-theoretic semantics as a starting point,
defining meaning in terms of \textit{truth}.
Given an \textit{individual} (also called an \textit{entity}),
and given a \textit{predicate} (the meaning of a content word),
we can ask whether the predicate is true of that individual.
Note that an individual could be
a person, an object, or an \textit{event},
following neo-Davidsonian event semantics \citep{davidson1967event,parsons1990event}.

Functional Distributional Semantics therefore
separates the modeling of words and individuals.
An individual is represented in a high-dimensional feature space.
The term \textit{pixie} refers to the representation of an individual \citep{emerson-copestake-2017-semantic}.
A predicate is formalized as a binary classifier over pixies.
It assigns the value \textit{true} if an individual with those features
could be described by the predicate,
and it assigns false otherwise.
Such a classifier is called a \textit{semantic function}.


The model is separated into a \textit{world model} and a \textit{lexicon model}.
The lexicon model consists of semantic functions.
Following situation semantics \citep{barwise1983situation},
the world model defines a distribution over \textit{situations}.
Each situation consists of a set of individuals, connected by semantic roles.
In our work, we only consider two types of semantic roles: ARG1 and ARG2.
For example, the sentence `a computer is on a desk'
describes a situation with three individuals:
the computer, the desk, and the event of the computer being on the desk.
The computer is the ARG1 of the event, and the desk is the ARG2, as shown in Figs. \ref{example} and~\ref{generative}.

Unlike other distributional models,
Functional Distributional Semantics is interpretable in formal semantic terms,
and supports first-order logic \citep{emerson2020quant}.
\citet{emerson2020autoencoding} proposed an autoencoder-like structure
which can be trained efficiently from semantic dependency graphs.

Because individuals are explicitly modeled,
grounding the pixies is more theoretically sound than grounding word vectors.
The framework has clear potential for learning grounded semantics,
which we explore in this paper.

\subsection{Visual Genome}
\label{sec:vg}

\begin{figure}[t]
\begin{center}%
\includegraphics[trim=0 29 0 24, clip, width=\linewidth]{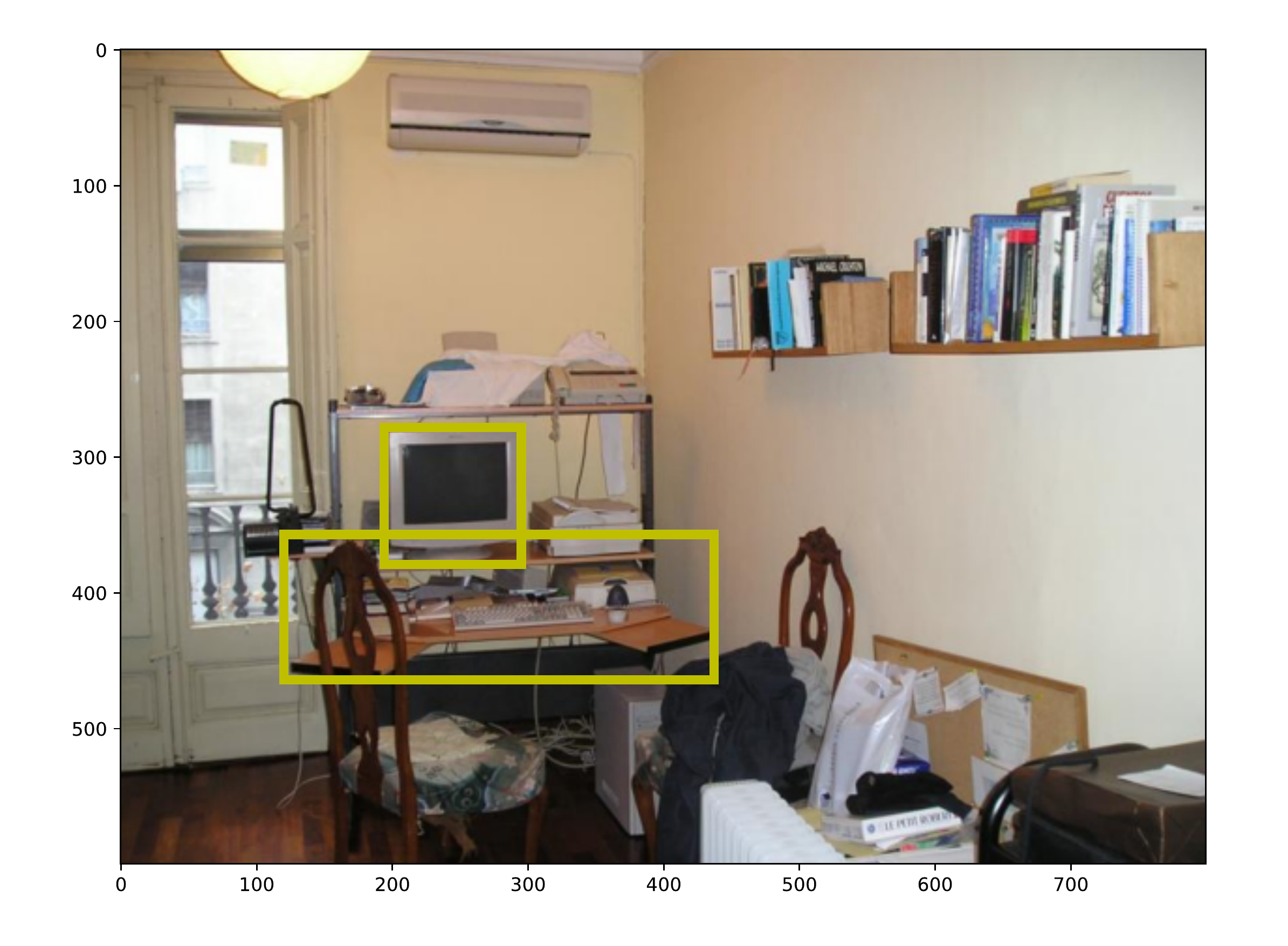}%
\vspace*{-1mm}%
\caption{An example image in Visual Genome,
annotated with the relation [`Computer', `ON', `Desk']}
\label{example}
\end{center}
\end{figure}

The Visual Genome dataset contains over 108,000 images and five different formats of annotations, including regions, attributes, relations, object instances and question answering.
In this work, we only consider the relations, which are formulated as predicate triples.
Each triple contains two objects in the image and one relation between them. The objects are identified with bounding boxes, as illustrated 
in Fig.~\ref{example}.
The object predicates are nouns or noun phrases,
and the relation predicates are verbs, prepositions or prepositional phrases.

Many works use Visual Genome as a grounded data source. For example, \citet{fukui-etal-2016-multimodal} use it to ground its visual question answering system.
Furthermore, the fine-grained annotations make Visual Genome a compelling dataset for studying lexical semantics.
As discussed by \citet{herbelot-2020-solve}, Visual Genome is similar in size to what a young child is exposed to, and the annotations are similar to simple utterances encountered during early language acquisition. 
\citet{kuzmenko-herbelot-2019-distributional} and \citet{herbelot-2020-solve}
learn semantics from the annotations, while discarding the images themselves.
They trained word embeddings with a count-based method and a Skip-gram-based method, respectively.
This methodology, of extracting word relations from an annotated image dataset, was also analyzed and justified by \citet{schlangen-2019-natural}.
In fact, \citet{verHo2021efficient} analyze the different modalities in Visual Genome in terms of information gain,
and conclude that, for enriching a textual model, the relational information provides more potential than the visual information.

To our knowledge, there has been no previous attempt to use grounded visual data to train a Functional Distributional Semantics model, nor to utilize the visual information of Visual Genome to learn natural language semantics.

\section{Model and Methods}
\label{sec:model}
We will explain the probabilistic structure of our model in Section~\ref{graphical}, and how we train the components in Sections \ref{world} and~\ref{lexicon}. 
In Section~\ref{VI}, we present an inference model to infer latent pixies from words and the context. 
\subsection{Probabilistic Graphical Model}
\label{graphical}

\begin{figure}
\begin{center}
\includegraphics[ width=\linewidth]{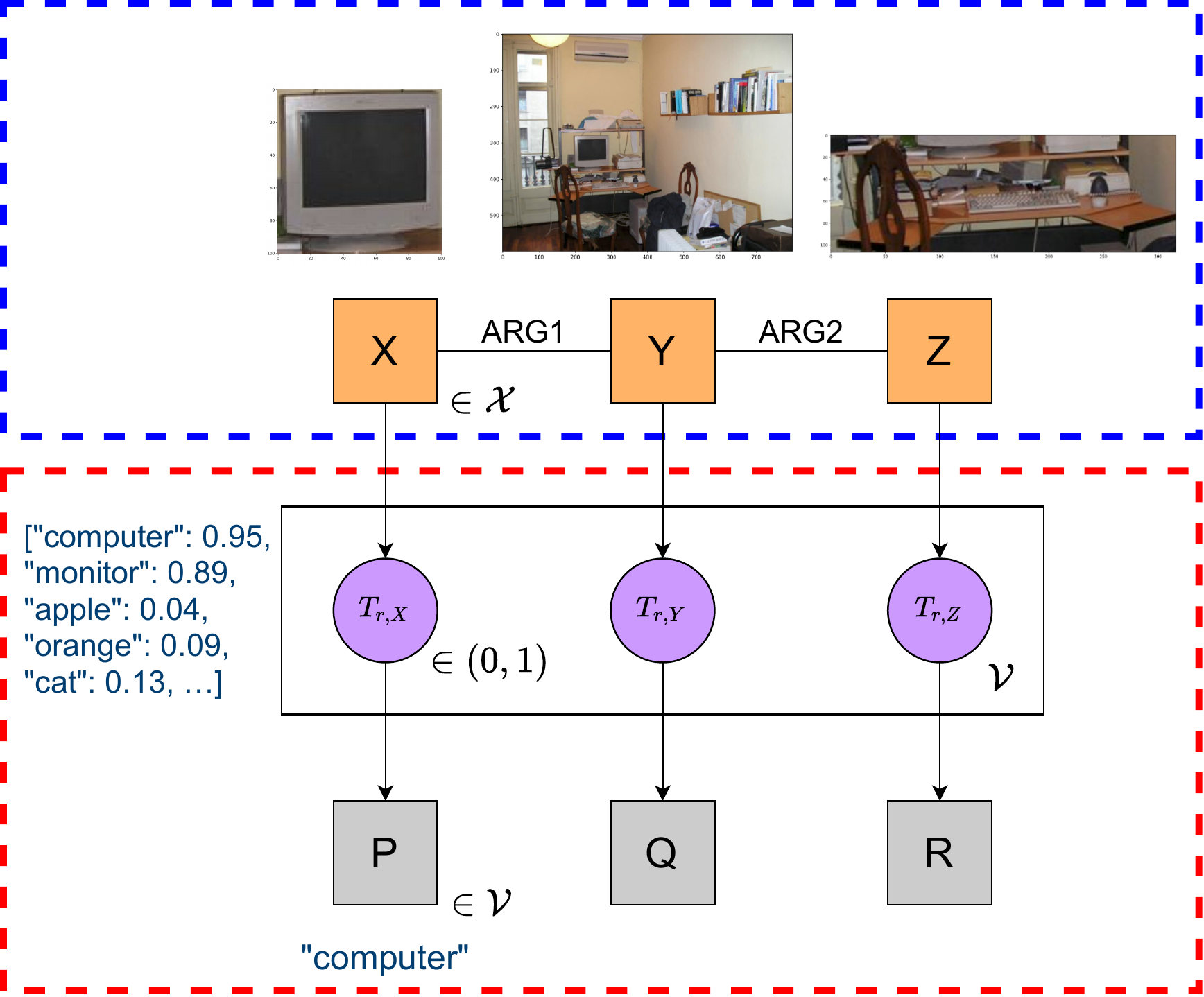}
\caption{Our probabilistic graphical model. 
The top blue box contains the world model, which learns the joint distribution of the observed pixies $X$, $Y$ and $Z$ from their corresponding images. The bottom red box shows the lexicon model, where each semantic function in the vocabulary~$\mathcal{V}$ is applied to each pixie.
For each pixie, one predicate is generated,
with probability proportional to the probability of truth.}
\label{generative}
\end{center}
\end{figure}

We define a graphical model which jointly generates pixies and predicates, as shown in Fig.~\ref{generative}.
It has two parts.
The world model is shown in the top blue box, which models the distribution of situations, or in other words, the joint distribution of pixies.
It is an undirected graphical model, with probabilistic dependence according to the ARG1 and ARG2 roles,
as further explained in Section~\ref{world}.
The lexicon model is shown in the bottom red box, which models each predicate as a semantic function.
It is a directed graphical model.
For each pixie, it produces
a probability of truth for each predicate (which are not observed),
as well as generating a single predicate (which is observed),
as further explained in Section~\ref{lexicon}.
Our framework can perform contextualized inference of predicate triples, where the world model provides contextual dependency while the lexicon model focuses on individual predicate, 
as further expalined in Section~\ref{VI}.

Given a labeled image triple, the model can be trained by maximizing the likelihood of generating the data, including both observed predicates and observed pixies. 
The likelihood can be split into two parts, as shown in Eq.~\ref{gradient}, where $s$ is a situation (a pixie for each individual), and $g$ is a semantic dependency graph (a predicate for each individual).
The first term is the likelihood of generating the observed situation, modeled by the world model. The second term is the likelihood of generating the dependency graph given an observed situation, modeled by the lexicon model. 
Therefore, we can optimize parameters of the two parts separately.
\vspace*{-1mm}
\begin{equation} 
    \log P(s,g) = \log P(s) + \log P(g|s)
    \label{gradient}
\end{equation}

\subsection{World Model}
\label{world}

The world model learns the joint distribution of pixies, as shown in the top half of Fig.~\ref{generative}.
The individuals are grounded by images, so we can obtain the pixie vectors by extracting visual features for individuals from their corresponding images.
For object pixies, they are grounded by their corresponding bounding boxes.
For event pixies, Visual Genome does not have labeled bounding boxes for them and their meaning tends to be more abstract, so we use the whole image to ground them.
As a feature extractor, we use ResNet101, a Convolutional Neural Network (CNN) pre-trained on ImageNet.
To further reduce redundant dimensions, we perform PCA on the last layer of the CNN. 
We take the output of PCA as the pixie space $\mathcal{X}$.
A situation $s$ is a collection of pixies within a semantic graph.
In this work, we only consider graphs with three nodes,
connected by the roles ARG1 and ARG2,
to match the structure of Visual Genome relations.

In previous work, the world model was implemented as a Restricted Boltzmann Machine (RBM).
However, an RBM uses binary-valued vectors,
which is not compatible with the real-valued vectors produced by a CNN.
Furthermore, an RBM does not give normalized probabilities,
which means that computationally expensive techniques are required,
such as MCMC, used by \citep{emerson-copestake-2016-functional},
or Belief Propagation, used by \citep{emerson2020autoencoding}.

We model situations with a Gaussian Markov Random Field (MRF).
For an $n$-dimensional pixie space, this gives a $3n$-dimensional Gaussian distribution,
with parameters $\mu$ and $\Sigma$ for the mean and covariance.
As shown in the first term of Eq.~\ref{gradient},
we would like to maximize $P(s)$.
\vspace*{-1mm}
\begin{equation}
    P(s) = \mathcal{N}(s;\mu, \Sigma)
\end{equation}

For a Gaussian distribution, the maximum likelihood estimate (MLE) has a closed-form solution,
which is simply the sample mean and sample covariance.
However, because we assume the left and right pixies in Fig.~\ref{generative} are conditionally independent given the event pixie, we force the top right and bottom left pixie blocks of the precision matrix $\Sigma^{-1}$ to be zero. We raise this assumption for the consideration of applying the Functional Distributional Semantics model to larger graphs with more individuals in the future. 
The assumption does not affect performance
on word similarity datasets,
but it slightly damages performance on contextual inference datasets.
Detailed results and discussion are given in Appendix~\ref{sec:CI}.


\subsection{Lexicon Model}
\label{lexicon}

The lexicon model learns a list of semantic functions, each corresponds to a word in predicate vocabulary $\mathcal{V}$. 
The semantic function $t_{r}(x)$ for a given predicate $r$ is a logistic regression classifier over the pixie space,
with a weight vector~$v_{r}$.
From the perspective of deep learning, 
this is a single neural net layer with a sigmoid activation function. As shown in Eq.~\ref{predicate}, the output is a probabilistic truth value ranging between $(0,1)$.
\vspace*{-1mm}
\begin{equation} \label{predicate}
t_{r}(x) = \sigma(v_{r}\cdot x) 
\end{equation}

As shown in the second row of Fig.~\ref{generative},
all semantic functions are applied to each pixie.
Based on the probabilities of truth, a single predicate is generated.
The probability of generating a specific predicate $r$ for a given pixie $\mathbf{x}$ is computed as shown in Eq.~\ref{pred_prob}. The more likely a predicate is to be true, the more likely it is to be generated.
\vspace*{-1mm}
\begin{align}\label{pred_prob}
    P(r|x) = \frac{t_{r}(x)}{\sum_{i} t_{i}(x)} 
\end{align}

The lexicon model is optimized to maximize $\log P(g|s)$, the log-likelihood of generating the predicates given the pixies.
This can be done by gradient descent.

\subsection{Variational Inference}\label{VI}
When learning from Visual Genome, pixies are grounded by images. However, when applying the model to text, the pixies are latent. 
We provide an inference model to infer latent pixie distributions given observed predicates. This inference model is used in Section~\ref{sec:extrinsic} on textual evaluation datasets.

Exact inference of the posterior $P(s|g)$ is intractable, because this requires integrating over the high-dimensional latent space of $s$. This is a common problem when working with probabilistic models. Therefore we use a variational inference algorithm to approximate the posterior distribution $P(s|g)$ with a Gaussian distribution $Q(s)$. 
For simplicity,
we assume that each dimension of $Q(s)$ is independent, so its covariance matrix is diagonal.

\begin{figure}[t]
\begin{center}
\includegraphics[trim=0 0 35 0, clip, width=0.95\linewidth]{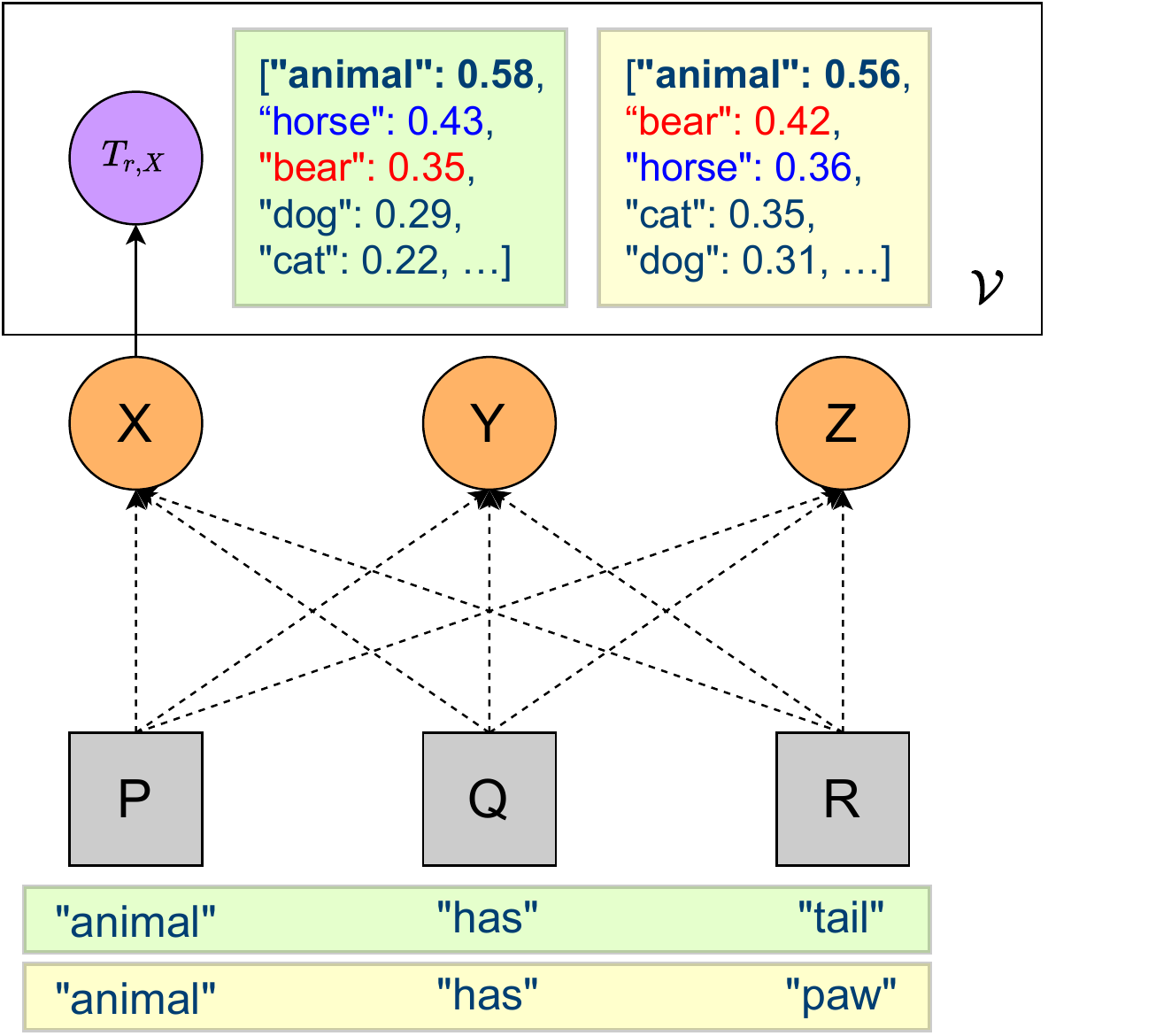}
\caption{Graphical inference model: The pixies $X$, $Y$ and $Z$ in the middle row are jointly inferred from the observed predicates $P$, $Q$ and $R$ in the bottom row, using variational inference.
Semantic functions are applied to~$X$, to give probabilities of truth.
As well as the observed predicate `animal',
the model predicts that other predicates may also be true.
Two examples are given, in green and yellow,
showing how the predicted truth values for `tail' and `paw'
depend on all observed predicates.
All probability values are inferred from our trained model.
}
\label{infergraph}
\end{center}
\vspace{-3mm}
\end{figure}

In Fig.~\ref{infergraph}, the graphical model illustrates this assumption, as there is no connection among the pixie nodes in the middle row.
Following the procedure of variational inference, the approximate distribution $Q(s)$ is optimized to maximize the Evidence Lower Bound (ELBO), given in Eq.~\ref{elbo}.
This can be done by gradient descent.
\vspace*{-1mm}
\begin{align} \label{elbo}
    \!\!\mathcal{L} 
    &= E_Q\Big[\log P(g|s) \Big] - \beta D_{\mathrm{KL}}\big(Q(s)||P(s)\big) 
\end{align}

The first term measures how well $Q(s)$ matches the observed predicates, according to the lexicon model $P(g|s)$.
The second term measures how well $Q(s)$ matches the world model $P(s)$. 
We would like to emphasize the likelihood of generating the observed predicates, so we down-weight the second term with a hyper-parameter $\beta$, similarly to a $\beta$-VAE \citep{Higgins2017betaVAELB}. Detailed analysis on the effects of $\beta$ is discussed in Appendix~\ref{sec:beta}.

Exactly computing the first term is intractable.
\citet{emerson2020autoencoding} used a probit approximation,
but we instead follow \citet{daunizeau2017semianalytical},
who derived the more accurate approximations given in Eqs. \ref{approx1} and~\ref{approx2},
where $x$ has mean $\mu$ and variance $\Sigma$.
The second approximation is particularly important,
as we aim to maximize the log-likelihood.%
\vspace*{-1mm}%
\begin{align} \label{approx1}
    E[\sigma(x)] &\approx \sigma \Bigg(\frac{\mu}{\sqrt{1+0.368\Sigma}} \Bigg) \\\label{approx2}
      E[\log \sigma(x)]  &\approx \log \sigma \Bigg( \frac{\mu - 0.319 \Sigma^{0.781}}{\sqrt{1+0.205 \Sigma^{0.870}}} \Bigg) 
\end{align}

The second term of Eq.~\ref{elbo} is the Kullback-Leibler (KL) divergence between two Gaussians, which has the closed-form formula given in Eq.~\ref{equ:KL}, where $k$ is the total dimensionality.
\begin{equation}
\begin{split}
\!\!\!D_{\textrm{KL}}(Q||P) = \frac{1}{2} \Big[\!\log\! \frac{|\Sigma_P|}{|\Sigma_Q|} -k + \textrm{tr}(\Sigma_P^{-1} \Sigma_Q) \\
+(\mu_Q {-} \mu_P)^T\Sigma_P^{-1} (\mu_Q {-} \mu_P) \Big]
\label{equ:KL}
\end{split}
\end{equation}


As illustrated in Fig.~\ref{infergraph},
variational inference allows us to calculate quantities
such as the probability that an animal which has a tail is a horse.
To obtain the inferred distribution for a single pixie,
we need to marginalize the situation distribution~$Q(s)$.
From the independence assumption, 
this simply means taking the parameters for the desired pixie.
Then we can apply the semantic function for~$r$ on the inferred pixie~$x$,
as shown in Eq.~\ref{eqn:approxtruth},
which can be approximated using Eq.~\ref{approx1}.
\begin{equation}
t_{r}(x) \approx E_Q \big[ \sigma(v_{r}\cdot x)  \big]
\label{eqn:approxtruth}
\end{equation}

Although $Q(s)$ assumes independence,
its parameters are jointly inferred based on all predicates.
This is because the KL-divergence in Eq.~\ref{equ:KL}
depends on $\Sigma_P$, which is nonzero between each pair of pixies linked by a semantic role.

For example, in Fig.~\ref{infergraph},
the truth of `horse' for~$X$ depends on the observed predicate `tail' or `paw'.
This is not a direct dependence between words, but rather relies on three intermediate representations (the three pixies), all of which are expressed in terms of visual features.
The first term of the ELBO connects the semantic function for `tail' or `paw' to the variational parameters for $Z$.
The second term of the ELBO connects the variational parameters for $Z$ and $Y$ (based on the world model covariance for ARG2)
as well as $Y$ and $X$ (based on the world model covariance for ARG1).
Finally the semantic function for `horse' is applied to the variational distribution for $X$.

In this example, the model correctly infers that an animal with a tail is more likely to be a horse and an animal with paws is more likely to be a bear.
We notice that the truth values are generally low for all semantic functions. Even the highest truth is only around 0.58. This illustrates that the model is not very certain,
which might be expected since the model is performing inference on visual features, but the training image data is noisy. 

For some evaluation datasets, we need to perform inference given a single predicate.
This can be done by marginalizing the joint distribution.
Which pixie variable to choose, out of the three, should depend on the Part-Of-Speech (POS) of the word.
For nouns, the pixie node $X$ or $Z$ should be used, as a noun should play the role of ARG1 or ARG2.
For verbs and prepositions, the node $Y$ should be used, as they usually describe the relation.

\section{Evaluation}
\label{evaluation}
To train our model, we follow the same pre-processing and filtering of Visual Genome as \citet{herbelot-2020-solve}.
Details of pre-processing and hyper-parameters are given in the appendix.
 
\subsection{Intrinsic Evaluation}
\label{sec:intrinsic}
In this section, we examine whether a Gaussian MRF is a suitable choice for the world model, and whether the pixies in the pixie space are linearly separable such that the logistic semantic functions can successfully classify them.

\subsubsection{World Model Evaluation}
The world model learns a Gaussian distribution for the observed situations. In this section, we justify this choice 
by evaluating the fitting errors.

Fig.~\ref{graph:directworld} shows density histograms for two example pixie dimensions and their corresponding best-fit (MLE) Gaussian curves. 
The left histogram is an example for a majority of the pixie dimensions,
which is tightly matched by the best-fit Gaussian.
In other cases, as shown on the right, there are imbalanced tails and asymmetry. Despite their skewness and kurtosis, which make them look more like a Gamma distribution, they are still generally bell-shaped
and the departure is not so heavy. 

To quantify the errors, we measure the Wasserstein distance, the area of the histogram missing from the best-fit Gaussian.
Across all 100 pixie dimensions, the mean percentage missing is $7\%$ with a variance of $1\%$. 
A more flexible model might give better modeling performance, which could be a future improvement direction.
Nonetheless, we consider this level of error to be acceptably low.

\subsubsection{Lexicon Model Evaluation}
\label{evaluation:lexicon}
In this experiment, we investigate if our approach to model the semantic functions as logistic regression classifiers is suitable. 
In particular, a logistic regression classifier is a linear classifier, which means if the data is not linearly separable, it would have inferior performance.

\begin{figure} [t]
\begin{center}
\includegraphics[trim=18 0 55 15, clip, width=\linewidth]{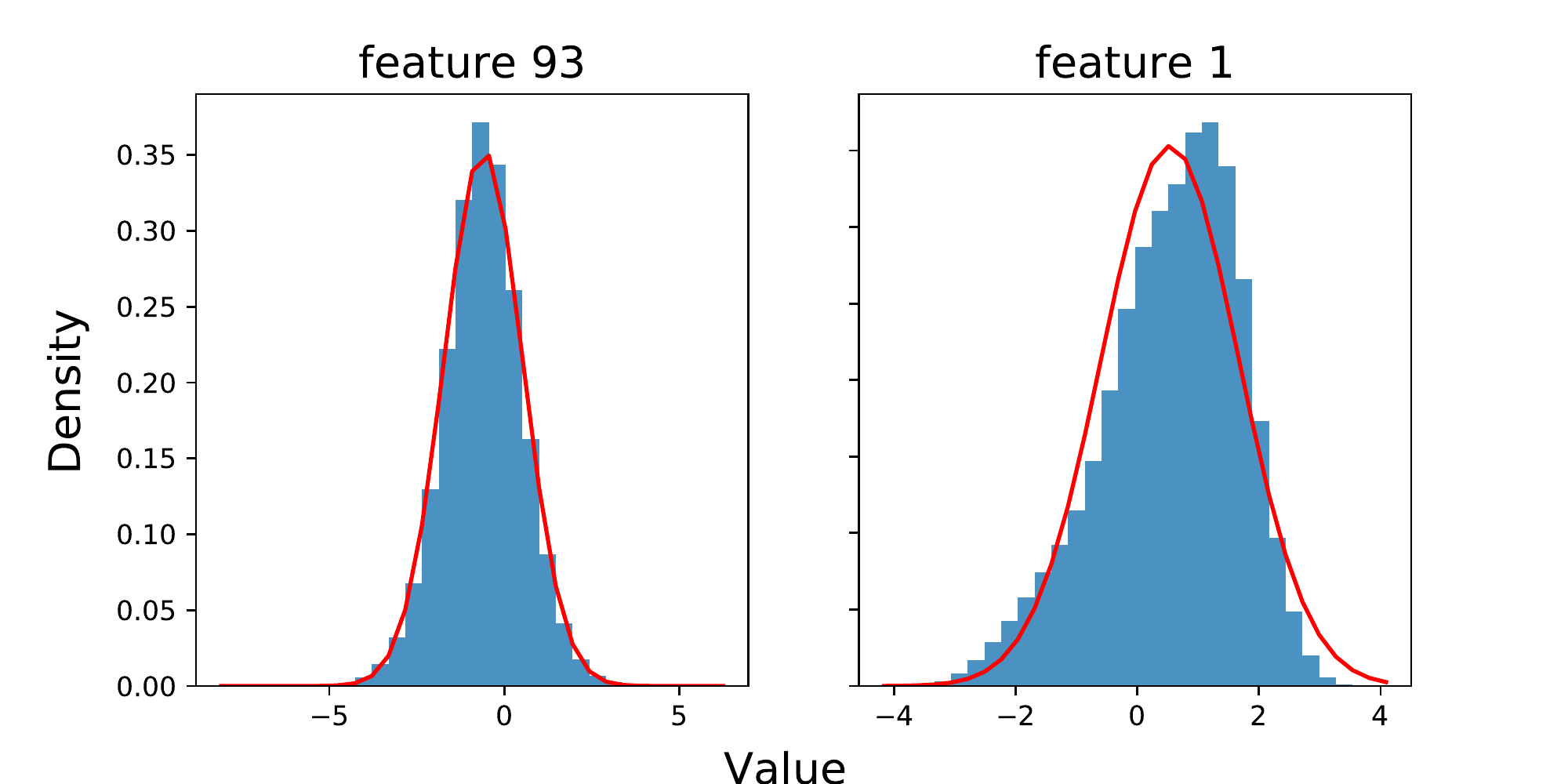}%
\vspace*{-2mm}%
\caption{
Density histograms for two selected pixie dimensions, across the 2.8M training instances. Best-fit Gaussian curves of the histograms are shown in red.}
\label{graph:directworld}
\includegraphics[trim=10 20 40 30, clip, width=0.95\linewidth]{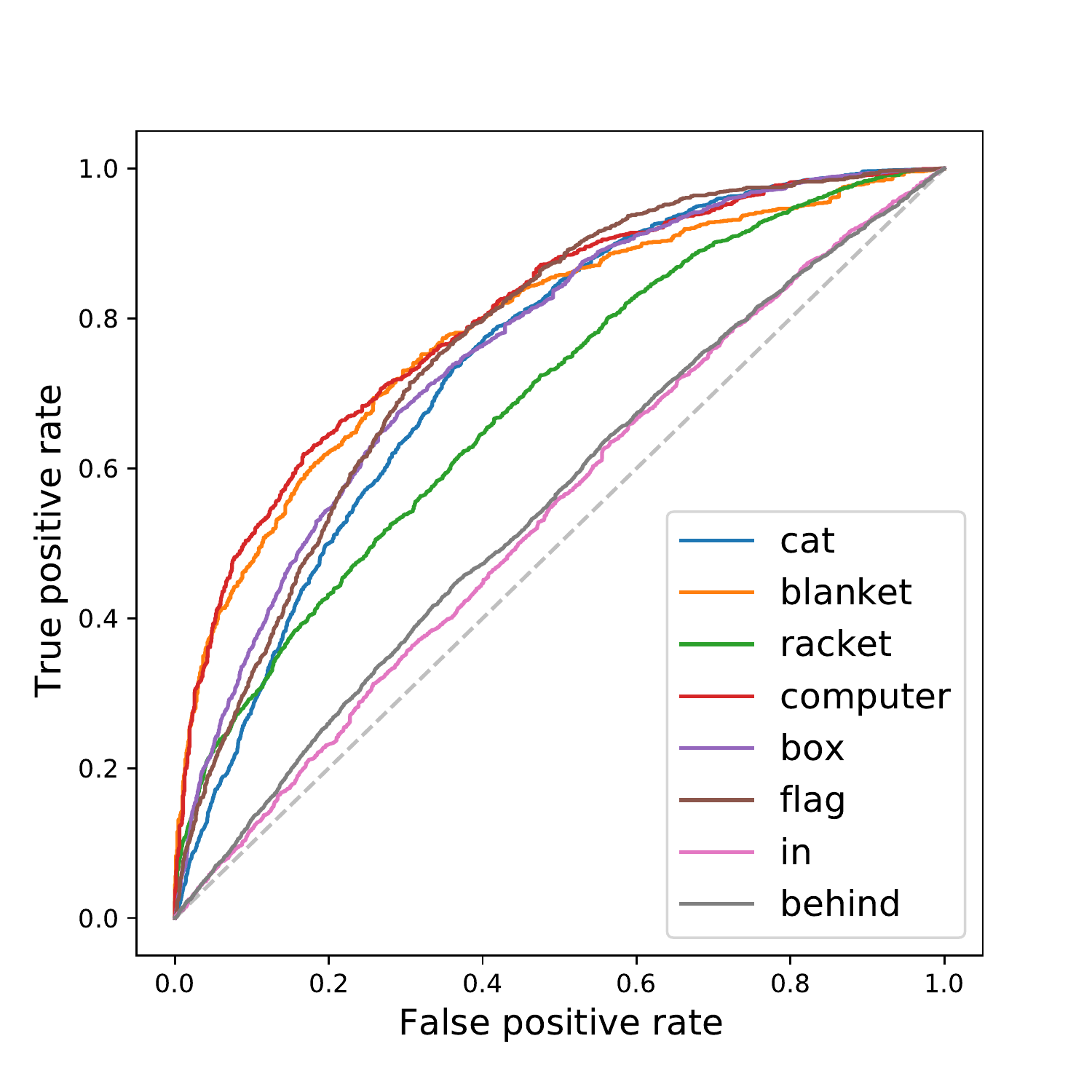}%
\vspace*{-2mm}%
\caption{ROC curves of the semantic functions for selected predicates.
Higher left is better performance.}
\label{roc}
\end{center}
\vspace{-1mm}
\end{figure}

We computed the Area Under Curve for the Receiver Operating Characteristic (AUC-ROC), for all predicates in the vocabulary.
For each predicate we randomly select equal amount of negative example pixies with its positive examples.
The average score is 0.79 for object predicates, and 0.58 for event predicates.
We also present the ROC for a few example predicates in Fig.~\ref{roc}.




We can see that object classifiers have generally better performance.
The classifier for `racket' shows slightly worse performance than the others, whose reason might be its lower frequency.
Compared to object predicates, the semantic functions for event predicates generally perform worse.  
There are two potential reasons which could be improved in future work.
Firstly, we used visual features generated from the whole image to represent the event pixie,
which is often not specific enough to identify the event.
Secondly, a logistic regression classifier might not be sophisticated enough for this classification problem.

\subsection{Extrinsic Evaluation}
\label{sec:extrinsic}

In this section, we use external semantic evaluation datasets, to give a direct comparison against previous work,
and to test whether our model can generalize beyond the training data. 
We evaluate on two lexical similarity datasets in Section~\ref{sec:lex-sim},
and two contextual datasets in Section~\ref{sec:context}.
We compare against two types of baseline: models trained on a large corpus and models trained on Visual Genome.

For these datasets, our model must assign similarity scores for predicate or triple pairs,
which we compute as follows.
The pixie values are inferred from the first predicate or triple in the pair. Then all semantic functions from the predicate vocabulary are applied to that pixie. Then the ranking of the second predicate in the pair over all potential predicates in the evaluation dataset is taken as the similarity score. Therefore, smaller ranking means higher similarity between predicates.

Finally, because there are discrepancies between vocabularies used in Visual Genome and the evaluation datasets,
we follow \citet{herbelot-2020-solve} in filtering the evaluation datasets according to the Visual Genome vocabulary, and use the filtered datasets to evaluate all models.
For the two lexical datasets,
we exactly follow \citeauthor{herbelot-2020-solve}'s filtering conditions to give a direct comparison.

For the contextual datasets, this filtering is too strict,
resulting in zero vocabulary coverage.
For these datasets, we apply looser filtering, with details given in the appendix.
This also requires re-training our model and the Visual Genome baselines on a more loosely filtered training set.

\begin{table*}[t]
\begin{center}
\begin{tabular}{l|l|c|c|c|c} 
\hhline{~=====}
                                       & \multirow{2}{*}{Models} & \multicolumn{2}{c|}{Lexical datasets~}   & \multicolumn{2}{c}{Contextual datasets}  \\ 
\cline{3-6}
\begin{tabular}[c]{@{}l@{}}\\\end{tabular}                                       &                         & ~~\textbf{MEN}~~ & ~\textbf{SimLex-999}~ & ~~\textbf{GS2011}~ & ~\textbf{RELPRON}~  \\ 
\hline
\multirow{3}{*}{\begin{tabular}[c]{@{}l@{}}Large corpus\\baselines\end{tabular}} & Word2vec-1B             & 0.641            & 0.384                 & 0.265              & 0.381               \\
                                                                                 & Word2vec-6B             & 0.652            & 0.397                 & 0.278              & 0.401               \\
                                                                                 & Glove-6B                & \textbf{0.717}   & 0.409                 & \textbf{0.293}     & \textbf{0.432}      \\ 
\hline
\multirow{3}{*}{VG baselines}                                                    & VG-count                & 0.336            & 0.224                 & 0.063              & 0.038               \\
                                                                                 & VG-retrieval            & 0.420            & 0.190                 & 0.072              & 0.045               \\
                                                                                 & EVA                     & 0.543            & 0.390                 & 0.068              & 0.032               \\ 
\hline
Proposed approach                                                                & Our model               & 0.639            & \textbf{0.431}        & 0.171              & 0.117               \\
\hline\hline
\end{tabular}
\caption{Evaluation results. For MEN, SimLex-999 and GS2011, the metric is Spearman correlation; for RELPRON, mean average precision. All models are evaluated on subsets of the data covered by the VG vocabulary.
}
\label{results}
\end{center}
\vspace{-1mm}
\end{table*}

\subsubsection{Baselines}

\textbf{Visual Genome Baselines}:
We re-implement two previously proposed models
learning distributional semantics from Visual Genome,
described in Section~\ref{sec:vg}.
A simple count-based model was proposed by \citet{kuzmenko-herbelot-2019-distributional},
which we refer to as VG-count.
\citet{herbelot-2020-solve} improved on this and proposed EVA, a Skip-gram model trained on the same kind of co-occurrence data.

We also implement an image-retrieval baseline which we refer to as VG-retrieval. 
This baseline simply retrieves all image boxes whose annotations match the indexing predicate. Visual features are extracted in the same method as our model, as described in \ref{world},
and then averaged across all retrieved images to obtain the representation for a given predicate. This baseline illustrates the performance can be achieved when only using the visual information of Visual Genome.



\noindent
\textbf{Large Corpus Baselines}:
We trained two Skip-gram Word2vec models \citep{mikolov2013efficient} using 1 billion and 6 billion tokens from Wikipedia, using Gensim \citep{Rehurek2010SoftwareFF}. We will refer to them as Word2vec-1B and Word2vec-6B. The window sizes are set to be 10 in two directions, so they contextualize with far more words than our model.
We also use Glove \citep{pennington-etal-2014-glove} trained on 6 billion Wikipedia tokens as another strong baseline, which we refer to as Glove-6B. For all three baselines, the dimensionality is set to 300.

Compared to the large corpus baselines,
our model has fewer parameters per word (100 vs.\ 300),
and is trained on far fewer data points (2.8M relation triples vs.\ 1B or 6B tokens).



\subsubsection{Lexical similarity and relatedness}
\label{sec:lex-sim}
We use two lexical similarity/relatedness datasets,
MEN \citep{men2014}
and Simlex-999 \citep{hill-etal-2015-simlex},
both of which give scores for pairs of words.
MEN contains 3000 word pairs,
and SimLex-999 contains 999 pairs.
After filtering for the Visual Genome vocabulary,
we have 584 pairs for MEN
and 169 pairs for SimLex-999.

MEN evaluates relatedness,
while SimLex-999 evaluates similarity.
For example, `coffee' and `cup' are related, but not similar.
Capturing similarity rather than relatedness is hard for most text-based distributional semantics models because they build concept representations based on their co-occurrence in corpora, which generally reflects relatedness but not similarity. However, similarity might be more directly reflected in terms of visual features which can be captured by our model.

The results are shown in Tab.~\ref{results}.
Our model
outperforms the two baselines trained on Visual Genome, and matched the performance of Word2vec-1B
(the difference is statistically insignificant, $p{>}0.5$).

If we force our model to evaluate on the full 1000 word pairs in the MEN test set (assigning the median similarity score to the out-of-vocabulary pairs), it still achieves 0.304. 
Using the loosely filtered training set, our model can achieve the even higher score of 0.670 (on the same strictly filtered subset of MEN).
This illustrates that one limit of our model's performance is the size of the Visual Genome dataset.
In contrast, the performance of Word2vec does not improve much as the training data increases from 1B to 6B, which suggests there is a limit on how much can be learnt from local textual co-occurrence information alone.

On SimLex-999, our model achieves 0.431, which outperforms all baselines.
Compared to Glove-6B, the strongest baseline,
it is weakly significant ($p{<}0.15$).
This might justify our point that there is advantage of learning similarity from visual features. Additionally, our model can use parameters and data more effectively and efficiently than Word2vec and Glove, 
achieving better performance with less training data and fewer parameters.

Compared with VG-count and EVA, our model
can understand more semantics because it
learns from the visual information.
While compared with VG-retrieval, our model 
can leverage textual co-occurrence.
As far as we know, we have achieved a new state of the art on learning lexical semantics from Visual Genome.
Combining results across all four datasets (including the contextual datasets below),
the difference between our model and EVA is highly significant ($p{<}0.001$).

\subsubsection{Contextual semantics}
\label{sec:context}

We consider two contextual evaluation datasets.
GS2011 \citep{grefenstette-sadrzadeh-2011-experimental}
gives similarities of verbs in a given context.
Each data point is a pair of subject-verb-object triples, where only the verbs are different.
For example, 
[`table’,`show’, `result’] and [`table’, `express’, `result’]
are judged highly similar.
The dataset has 199 distinct triple pairs and 2500 judgment records from different annotators.
The evaluation metric is Spearman correlation across all judgments.
As \citet{van-de-cruys-etal-2013-tensor} point out, the second verb in each pair is often nonsensical when combined with the corresponding subject and object. Therefore,
we only compare the triple pairs in a single direction, inferring pixies from the first triple and applying the second verb's semantic function.

RELPRON \citep{rimell-etal-2016-relpron}
evaluates compositional semantics.
It contains a list of terms, each associated with around 10 properties.
Each property is a noun modified by a subject or object relative clause.
For example, the term `theater' has the subject property [`building', `show', `film'] and object property [`audience', `exit', `building'].
The task is to find the correct properties for each term,
evaluated as Mean Average Precision (MAP).
The development set contains 65 terms and 518 properties; the test set, 73 terms and 569 properties.

Under the loosely filtered condition,
our subset of GS2011 contains 252 similarity judgments;
RELPRON, 57 terms and 150 properties.

\citet{rimell-etal-2016-relpron} find that vector addition performs surprisingly well at combining contextual information. Therefore, for all baselines, we
represent a triple by taking the addition of the three word representations.
As aforementioned, we re-train our model and the VG baselines with loosely filtered data.

The results are shown in Tab.~\ref{results}. 
The corpus models outperform the VG models.
However, this is perhaps expected given that the vocabulary in GS2011 and RELPRON is more formal, 
and even when they are covered in Visual Genome, their 
frequencies are low:
for RELPRON, 54\% of the covered vocabulary has frequency below 100, compared to only 6\% for MEN.
Furthermore, GS2011 evaluates similarity of verbs,
but we saw in Section~\ref{evaluation:lexicon}
that our model is less accurate for verbs.

However, our model outperforms all VG baselines on both datasets.
This suggests that our model is less affected by data sparsity.
For the baselines, if a training triple contains multiple rare predicates, the sparsity problem is compounded.
However, our model relies on the images,
whose visual features are shared across the whole training set.



\subsection{Truth regularization}
\label{sec:truth-reg}

To make the probabilistic truth values more interpretable,
\citet{emerson2020autoencoding} proposes a regularization term
which penalizes the model if all truth values stay close to~0.
This would modify the loss function in Eq.~\ref{pred_prob},
to give Eq.~\ref{eqn:truth-reg},
with a hyper-parameter $\alpha$ that we set to 0.5.
\begin{equation}
\mathcal{L} = \log\frac{t_r(x)}{\sum_i t_i(x)} + \alpha\log t_r(x)
\label{eqn:truth-reg}    
\end{equation}%



\begin{figure} [t]
\begin{center}
\includegraphics[trim=10 20 40 50, clip, width=\linewidth]{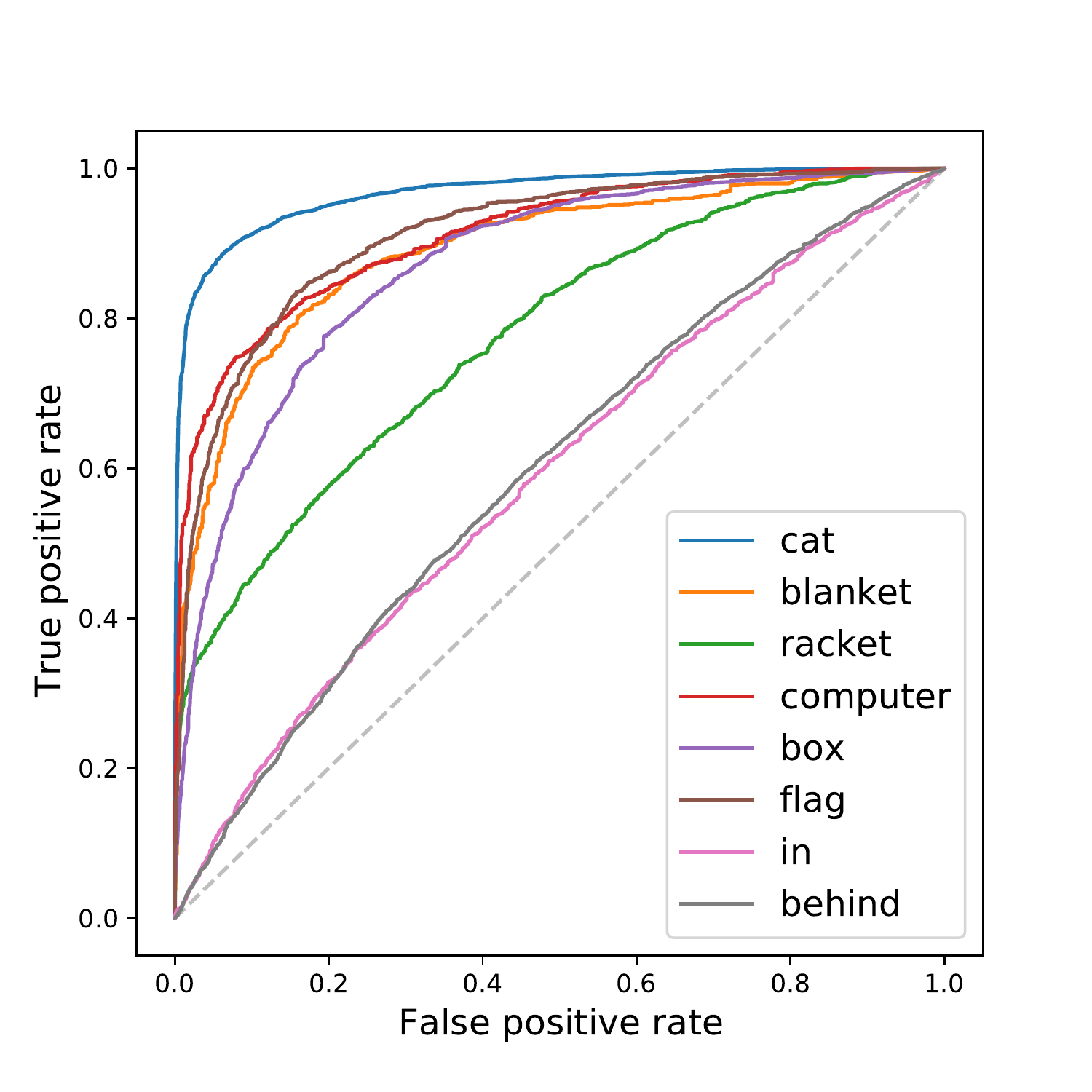}
\caption{ROC curves of the semantic functions for selected predicates,
for the truth-regularized model.
}
\label{roc2}
\end{center}
\end{figure}

We find that adding the log-truth term
improves performance on intrinsic evaluation,
but decreases performance on extrinsic evaluation.
Applying the analysis in Section~\ref{evaluation:lexicon},
the average AUC-ROC is 0.86 for object predicates and 0.60 for event predicates.
This is illustrated in Fig.~\ref{roc2} for the same example predicates as Fig.~\ref{roc}.
In contrast, when evaluating on MEN and SimLex-999,
this model achieves only 0.602 and 0.381 respectively.
On GS2011 and RELPRON, the model achieves lower performance of 0.112 and 0.056. 

The log-truth term makes predicates true over larger regions of pixie space.
As shown by the intrinsic evaluation, this is helpful when considering each classifier individually.
However, the regions of different predicates also overlap more, which seems to hurt their overall performance on the external datasets.
To quantify this, 
we calculate the total truth of all predicates, for 1000 randomly selected images. For the original version of our model, on average 0.83 predicates are true for an image.
This is slightly below~1, illustrating the problem
\citeauthor{emerson2020autoencoding} aimed to avoid.
However, with the log-truth term, it becomes 25.5,
which may have over-corrected the problem.

\section{Conclusion}

In this paper, we proposed a method to train a Functional Distributional Semantics model with visual data. 
Our model outperformed the previous works and achieved a new state of the art on learning natural language semantics from Visual Genome. 
Further to this, our model achieved better performance than Word2vec and Glove on Simlex-999 and matched Word2vec-1B on MEN.
This shows that our model can use parameters and data more efficiently than Word2vec and Glove.
Additionally, we also showed that our model can
successfully be used to make contextual inferences.
As future work, we could leverage previous work to jointly train the Functional Distributional Semantics model with both visual and textual data, such that we could improve the vocabulary coverage and have better understanding of abstract words.

\bibliography{anthology,acl2020}
\bibliographystyle{acl_natbib}

\appendix

\section{Training and evaluation details}
\label{training}

\subsection{Filtering}
For EVA, \citet{herbelot-2020-solve} filtered the Visual Genome dataset with a minimum occurrence frequency threshold of 100 in both ARG1 and ARG2 directions. After filtering, the resulting subset contains 2.8M relation triples and the vocabulary size is 1595.
When evaluating the external datasets, it only includes the noun predicates.
For the results reported in the intrinsic evaluations and in Tab.~\ref{results} where we specify `strict filtering', we follow the same filtering conditions with EVA. 

We also train our model with a less strict filtering setting, where the minimum frequency threshold is set at 10 in at least one direction. Under this filtering setting, the resulting subset contains 3.4M relation triples and the vocabulary size is 6788.
When evaluating the external datasets, we includes all the covered predicates regardless of their POS. 
The results under the `loose filtering' columns in Tab.~\ref{results} are evaluating under this setting. Additionally, every time we use the model trained under this setting, we will emphasize it is `under less strict filtering condition'.

\subsection{CNN}
\label{sec:CNN}
For the visual feature extractor, the pretrained CNN, we used ResNet101 \citep{he2016deep}, which has 101 layers deep and trained on ImageNet \citep{deng2009imagenet}.

\subsection{PCA}
During the PCA transform, we reduce the pixie dimension from 1000 to 100, whose eigenvalue components cover $93.2\%$ of the total information.
After the PCA, we re-scaled each dimension by dividing them over the square root of their corresponding eigenvalues and scale up by a factor of 1.15, such that the determinant of the covariance matrix of the world model is close to 1.

\subsection{Conditional independence assumption}
\label{sec:CI}
The conditional independence assumption of the world model is raised for the consideration of applying our framework to larger graphs with more individuals in the future. It allows us to decompose a complicated graph in terms of relations. Otherwise, a separate precision matrix is required for each graph topology.  
However, this assumption theoretically damages the ability of contextual inference of our model,
as pixies $X$ and $Z$ are only dependent on one another
via the pixie $Y$.

To investigate the effects of this assumption quantitatively, we performed experiment to compare the evaluation results under two settings. All other settings of hyperparameters remain the same. 
For single-word similarity datasets MEN and SimLex-999,
the effect on performance is inconsistent,
and the differences are statistically insignificant ($p{>}0.5$).
For contextual datasets GS2011 and RELPRON, releasing the assumption improves results, and the differences are statistically significant ($p{<}0.1$ for each dataset, and $p{<}0.01$ when combining both datasets).

Possible avenues for future work would be to improve the modeling of events, which could make the conditional independence assumption more reasonable
(recall that in Section~\ref{sec:intrinsic}, the modeling of events was identified as a limitation),
or to modify the graphical model to make it more flexible (which would be a challenge for larger graphs).


\begin{table}[t]
\begin{center}
\begin{tabular}{l|l|l|}
\cline{2-3}
                                 & With CI & Without CI \\ \hline
\multicolumn{1}{|l|}{MEN}        & 0.639   & 0.658      \\ \hline
\multicolumn{1}{|l|}{SimLex-999} & 0.430   & 0.410      \\ \hline
\multicolumn{1}{|l|}{GS2011}     & 0.171   & 0.182      \\ \hline
\multicolumn{1}{|l|}{RELPRON}    & 0.117   & 0.137      \\ \hline
\end{tabular}
\end{center}
\caption{Evaluation results. For MEN, SimLex-999 and GS2011, the metric is Spearman correlation; for RELPRON, mean average precision. All models are evaluated on subsets of the data covered by the VG vocabulary.}
\label{tab:CI}
\end{table}

\subsection{Lexicon model training}
When training the lexicon model, we used L2 regularisation with a weight of $5\mathrm{e}{-8}$ and the Adam optimizer \citep{kingma2015adam}. We train the lexicon model for 40 epochs and the learning rate is set at 0.01 with a step scheduler which reduces the learning rate by a factor of 0.4 every 5 epochs. 
The hyper-parameters are tuned on the training data
to maximize the number of predicates such that at least one image annotated with that predicate has a truth value of at least 0.1. For the model trained on strictly filtered data, the number of such predicates reaches 1343 out of the vocabulary size 1595, while for loosely filtered model, the number is 4453 out of 6788.

\subsection{Variational inference optimization}
For the variational inference, the hyper-parameter $\beta$ is set to be 0.1. We run gradient descent for 800 epochs with initial learning rate of 0.03 and a step scheduler which reduces the learning rate by a factor of 0.6 every 50 epochs.
The hyper-parameters for variational inference are tuned to maximize the ELBO on the filtered subset of MEN.
(The ELBO does not depend on the similarity scores,
just the input triples.)
Tuning these hyper-parameters has no effect on the training of the world model or the lexicon model.
The scores shown in Table~\ref{results} are the results averaged over 5 random seeds.

\subsection{Effects of hyperparameter \(\beta\)}
\label{sec:beta}
The hyperparameter \(\beta\) in ELBO, Equation~\ref{elbo}, controls the weighting of prior knowledge during inference. Higher \(\beta\) value will drag the inferred pixies closer to the average positions of all seen pixies with corresponding semantic roles. Lower  \(\beta\) will push the inferred pixies to the center of semantic functions of their corresponding predicates. 
In Table~\ref{tab:beta}, we show how the \(\beta\) affects the evaluation results. For single-word similarity dataset SimLex-999, better knowledge of the semantic functions can give more information than prior knowledge of average positions of their semantic roles. Therefore lower value is preferred. 
On contrast, for contextual dataset RELPRON, emphasizing the prior information could benefit the estimate of the jointly distributed pixie triples. The performance peaks at the value of 0.4.

\begin{table}[t]
\begin{center}
\begin{tabular}{|c|c|c|}
\hline
\(\beta\) & SimLex-999     & RELPRON        \\ \hline
0.05 & 0.395          & 0.091          \\ \hline
0.1  & \textbf{0.430} & 0.117          \\ \hline
0.2  & 0.35           & 0.123          \\ \hline
0.3  & 0.289          & 0.133 \\ \hline
0.4  & 0.194           & \textbf{0.150}         \\ \hline
0.5  & 0.083          & 0.144          \\ \hline
\end{tabular} 
\end{center}
\caption{Evaluation results against different \(\beta\). For SimLex-999 the metric is Spearman correlation and for RELPRON, mean average precision. All other settings hyperparameters remain the same.}
\label{tab:beta}
\end{table}

\subsection{Statistical tests}
All statistical tests are two-tailed bootstrap tests,
which follows the recommendations of \citet{dror-etal-2018-hitchhikers}.
We use 1000 samples for each test.

\end{document}